# Parking, Perception, and Retail: Street-Level Determinants of Community Vitality in Harbin

Haotian Lan

**Abstract:** The commercial vitality of community-scale streets in Chinese cities is shaped by complex interactions between vehicular accessibility, environmental quality, and pedestrian perception.  This study proposes an interpretable, image-based framework to examine how street-level features — including parked vehicle density, greenery, cleanliness, and street width—impact retail performance and user satisfaction in Harbin, China.  Leveraging street view imagery and a multimodal large language model (VisualGLM-6B), we construct a Community Commercial Vitality Index (CCVI) from Meituan and Dianping data and analyze its relationship with spatial attributes extracted via GPT-4–based perception modeling.

Our findings reveal that while moderate vehicle presence may enhance commercial access, excessive on-street parking — especially in narrow streets — erodes walkability and reduces both satisfaction and shop-level pricing.  In contrast, streets with higher perceived greenery and cleanliness show significantly greater satisfaction scores but only weak associations with pricing.  Street width moderates the effects of vehicle presence, underscoring the importance of spatial configuration.  These results demonstrate the value of integrating AI-assisted perception with urban morphological analysis to capture non-linear and context-sensitive drivers of commercial success.

This study advances both theoretical and methodological frontiers by highlighting the conditional role of vehicle activity in neighborhood commerce and demonstrating the feasibility of multimodal AI for perceptual urban diagnostics.  The implications extend to urban design, parking management, and scalable planning tools for community revitalization.

**Keywords:** Community commerce; Street view imagery; Multimodal large models (MLLMs); Parking intensity; Perceived streetscape quality

## 1.Introduction

The vitality of neighborhood-scale commercial zones — comprising shops, cafes, and services catering to local residents—is a key indicator of urban livability [1]. In increasingly auto-oriented cities, however, the growth of private vehicle use introduces both benefits and spatial tensions [2]. While cars can expand access and attract non-local customers, high vehicle density and on-street parking often compromise walkability and degrade the pedestrian environments that local commerce depends on [3].

This tension is particularly evident in Chinese cities such as Harbin, where car ownership has surged in recent decades [4]. Harbin's urban structure is dominated by gated residential communities, a development pattern that concentrates commercial activity along perimeter streets [5]. These boundary spaces, functioning as semi-public interfaces between enclosed living areas and the broader city, vary significantly in their atmosphere and commercial performance. Yet it remains unclear how the presence of motor vehicles — parked cars, curbside traffic, and parking infrastructure—shapes such differences in street vitality.

Although existing studies have explored how land use mix, density, and streetscape design influence commercial success [6], few have directly examined the micro-scale effects of vehicle presence on neighborhood commerce. Classic works such as Appleyard's Livable Streets and Jacobs's vision of sidewalk life emphasize the importance of walkable, socially vibrant streets [7][8]. However, in the context of semi-permeable Chinese communities, the role of vehicles in shaping commercial outcomes remains largely unexplored [5].

Harbin, a major city in Northeast China, provides a representative case for studying how motor vehicle presence affects neighborhood commercial vitality. Its combination of historic districts and gated communities presents varied street forms and parking conditions, while its cold climate intensifies seasonal reliance on automobiles [9]. These features make Harbin broadly reflective of Chinese cities facing growing motorization and evolving neighborhood structures.

This study investigates how the presence of motor vehicles—particularly parked vehicle density and parking infrastructure—affects community-scale commercial activity in Harbin. Using street-level imagery and multi-modal AI models, we extract measurable indicators of the physical street environment. These are integrated with fine-grained commercial data from platforms such as Meituan and Dazhong Dianping to assess the vitality of neighborhood businesses. Our analysis links macro-level attributes (e.g., location, accessibility) with micro-level streetscape features (e.g., greenery, street width, cleanliness) to explain variation in commercial performance.

We address three research questions: (1) How are parked vehicle density and parking availability associated with community commercial activity? (2) How do other physical street attributes—such as greenery, width, and cleanliness—compare in their influence? (3) What integrated framework can best explain variations in commercial vitality, accounting for both urban context and local street conditions?

By answering these questions, this paper contributes to the literature on urban commercial dynamics by highlighting the role of motor vehicles at the community scale and demonstrating the analytical value of AI-driven, image-based urban diagnostics. The remainder of the paper reviews relevant literature, outlines the research design, presents the results, discusses the findings, and concludes with policy recommendations.

## 2.Literature Review

*2.1.Reviewing the Evidence: Parking Infrastructure and Urban Retail Dynamics*

The relationship between motor vehicles, parking, and commercial vitality has long been debated in urban planning. A common belief among business owners is that easy car access and abundant parking are critical for retail success, and that pedestrianization or parking restrictions may negatively impact business performance. However, empirical studies offer a more nuanced understanding.

Recent research from European city centers suggests that excessive on-street parking can actually detract from retail performance, while proximity to off-street parking, pedestrian zones, and transit stops is positively associated with commercial success [10]. For instance, a 2023 study in Aachen, Germany found that a high density of curbside parking was correlated with lower retail rents, whereas nearby parking garages and walkable conditions supported higher commercial attractiveness [10]. These findings challenge the assumption that parking must be directly adjacent to shops. Instead, they suggest that accessible but well-planned parking — within walking distance — can free up valuable streetfront space for sidewalks, seating, or greenery, thereby enhancing the pedestrian experience [11].

In other contexts, such as the UK, surveys reveal that retailers tend to overestimate the proportion of customers who drive and underestimate the spending of those who walk or use transit [12]. These misperceptions further complicate assumptions about the role of cars in sustaining commercial vitality. Classic works like Appleyard's Livable Streets demonstrated that heavy traffic erodes social life on residential streets, reducing informal interaction and neighborhood cohesion [13]. This principle can be extended to commercial areas, where streets dominated by vehicles may discourage lingering, reduce foot traffic, and weaken customer engagement [14].

Conversely, a complete absence of parking or vehicular access may limit the catchment of commercial streets—especially in suburban or low-density areas. Urban vitality research therefore emphasizes balance: providing sufficient parking to maintain accessibility while preserving the walkability and social atmosphere of streets. The concept of "park-once environments" reflects this ideal, where visitors can leave their cars in a centralized facility and walk comfortably to multiple destinations [15].

In community-scale commercial settings, such as those in Harbin, this balance may be especially delicate. Local retail clusters often serve nearby residents but also attract some car-based visitors, particularly in wealthier communities. When on-street parking is poorly managed—either scarce or overly dominant—it can deter both pedestrian and vehicle-based users. This suggests a need for fine-grained studies that examine how different levels and configurations of vehicle presence influence commercial outcomes at the neighborhood scale [16-18].

*2.2. Perceptible Street Features and Commercial Vitality*

Beyond parking availability, a wide range of physical streetscape attributes influence pedestrian behavior and, consequently, commercial vitality. Among these, street width and traffic speed are particularly significant. Istrate found that on Shanghai residential streets, wider roadways showed a negative correlation with stationary pedestrian activity, despite national planning standards that often encourage wide carriageways [19]. Narrower streets, especially those where building facades closely align with sidewalks, promote walking and spontaneous interaction — benefiting cafes, small shops, and service-oriented retail. These observations echo long-standing principles in human-scale urban design, where disproportionally wide or fast-moving traffic corridors are associated with reduced walkability and diminished economic engagement [20, 21].

Streetscape greening is another critical determinant of street appeal. Multiple studies confirm that street trees and vegetation enhance not only aesthetic quality but also commercial outcomes. Trees provide shading, visual softness, and a more comfortable microclimate, increasing the likelihood that pedestrians will linger and engage with storefronts [22]. U.S.-based research has demonstrated that retail districts with high tree canopy coverage see higher footfall, longer dwell times, and even 9–12% higher customer willingness to pay for products [23]. Similar conclusions have been drawn in studies from Melbourne, Seoul, and Singapore, where tree-lined streets were rated more favorably in terms of perceived safety and ambience [24, 25]. In the context of Chinese cities, where street greening varies markedly by district and development age, indicators such as Green View Index (GVI) and visible planting are useful proxies for commercial street quality [26].

Cleanliness and physical maintenance also shape perceptions of safety, care, and professionalism—elements known to influence consumer behavior. Although rigorous causal studies on this topic remain limited, retail business analysts have long noted the direct benefits of clean sidewalks and storefronts, including improved visibility, stronger first impressions, and higher repeat visitation rates [27, 28]. In urban governance contexts, particularly within Chinese gated communities, the cleanliness of perimeter commercial streets often depends on property management standards or local district efforts, making it a key variable in explaining commercial performance variance across communities [29].

Land use diversity and frontage activity further contribute to commercial liveliness. Streets lined with a dense mix of small shops, visible displays, and frequent entrances tend to encourage pedestrian movement and spontaneous browsing. Istrate reports that streets with greater "permeability" and visible commercial variety support more stationary pedestrian behaviors and social interaction, especially when integrated with residential access points [19]. This aligns with the "active edge" theory of urban design, which posits that commercial success often correlates with continuous, engaging façades at eye level [30, 31]. In Chinese gated communities, however, commercial uses are typically confined to perimeter bands, limiting the natural integration of retail and residential space. Within this constraint, the spatial continuity, clustering intensity, and shopfront density of these bands may significantly influence customer flows.

*2.3. Street View Imagery and Multimodal AI in Urban Analysis*

Traditional studies of streetscape characteristics have long relied on field surveys, manual audits, or researcher-led observations. In recent years, however, the proliferation of street view imagery platforms (e.g., Google Street View, Baidu Maps) has transformed urban analysis, offering a scalable and standardized means to assess ground-level conditions across entire cities. According to a comprehensive review [32], street view imagery has rapidly become a critical geospatial data source in urban research, driven by advances in computer vision and machine learning.

A growing body of literature has leveraged such imagery to quantify urban design features—ranging from greenery and signage to façade articulation, sidewalk width, and even perceived safety or vibrancy [33–35]. This shift allows researchers to systematically extract micro-scale spatial features that are otherwise hard to detect through satellite images or conventional GIS datasets.

One major application involves computer vision–based object detection, where models such as YOLO, Faster R-CNN, or Mask R-CNN are trained to identify and count specific street elements. These models can, for example, detect vehicles, parking signage, crosswalks, or trees, enabling researchers to compute metrics like "vehicles per image" or "greenery ratio" [36, 37]. A commonly used metric, the Green View Index (GVI), calculates the proportion of green pixels in an image after semantic segmentation, serving as a proxy for street-level greenness [38].

More recently, the field has moved beyond feature detection to explore subjective perception modeling. Deep neural networks, especially convolutional neural networks (CNNs), have been trained on labeled street view datasets—often with human ratings on safety, comfort, or vibrancy—to predict perceptual qualities from new images [39]. This allows computational models to approximate how humans experience urban spaces, not just what physical elements exist. For example, one study combined image-derived features with point-of-interest (POI) density to predict urban vitality hotspots, showing that the inclusion of visual features significantly enhanced prediction accuracy [40].

A frontier development in this domain is the use of multimodal large language models (MLLMs), such as CogVLM or GPT-4V, which can process both image and text inputs. These models, pre-trained on vast image–text pairs, are capable of answering natural language queries about visual content (e.g., "How many cars are parked here?" or "Does this street appear clean?") [41, 42]. This enables more flexible and nuanced interpretation of streetscapes—capturing both quantitative (e.g., vehicle counts) and qualitative (e.g., perceived tidiness) aspects, which conventional computer vision may struggle to encode directly.

For example, one study analyzed spontaneous commercial setups in historic Chinese neighborhoods using over 10,000 street view images and AI-based workflows [43]. Their methodology integrated Mask R-CNN for object detection, random forest models for regression, and SHAP values for explainability, revealing strong links between informal vendor density and pedestrian activity. Their results support the potential of image-based metrics as robust proxies for commercial liveliness.

Inspired by such advances, our study combines image-derived features—such as parked vehicle density and visual greenery—with business activity indicators from digital commerce platforms. We also incorporate a multimodal foundation model (VisualGLM-6B), fine-tuned for street scene interpretation, making this one of the first applications of such models in a Chinese community-scale urban context. The inclusion of MLLMs enables us to extract both visual counts and qualitative descriptors (like perceived cleanliness), potentially offering a richer feature set for commercial vitality modeling.

In summary, three strands of literature inform our work:

(1) the complex and context-dependent relationship between vehicles/parking and commercial outcomes;

(2) the influence of diverse streetscape attributes on pedestrian and retail dynamics; and

(3) the emergence of AI-enabled methods for large-scale urban feature extraction using street imagery.

Our empirical study synthesizes these threads by applying state-of-the-art visual AI to street-level data from Harbin, examining how the micro-scale spatial environment correlates with the economic activity of community commercial areas.

## 3. Methodology

*3.1. Study Area and Units of Analysis*

This study examines Harbin, a major city in northeastern China with a metropolitan population exceeding five million. The city's residential fabric is largely composed of gated communities, each encompassing multiple apartment buildings enclosed within a defined perimeter. While internal areas of these communities typically lack commercial functions, small-scale retail and service activities are commonly distributed along the perimeter streets, forming linear commercial belts that cater primarily to local residents.

Our analysis focuses on these boundary-level commercial zones, which represent a key interface between the enclosed residential core and the public realm. Given the absence of commercial frontages in most community interiors, the study exclusively targets communities situated in peripheral districts where street-facing businesses are present and observable via panoramic imagery.

A sample of residential communities was selected to capture variation in spatial morphology, socio-economic status, and observed levels of commercial activity. Each community serves as the primary unit of analysis. For each unit, we derived commercial vitality indicators, quantified streetscape attributes such as vehicle presence and environmental conditions, and incorporated contextual urban variables extracted from both visual and geographic data. While the community is the core analytic scale, individual street segments along the community perimeter were also examined to support micro-spatial analysis where applicable.

3.2. Data Collection

*3.2.1 Street View Imagery Acquisition*

To extract physical characteristics of community commercial streetscapes, we utilized panoramic imagery from Baidu Street View, which provides high-resolution coverage for most urban streets in Harbin. As Google Street View is unavailable in China, Baidu served as the primary and consistent source.

Given our focus on summer conditions—when vegetation, pedestrian activity, and street cleanliness are most visible—we excluded any winter or snow-covered images from the dataset. Only summer imagery (2021–2023) was used to ensure comparability across scenes.

The sampling covered all accessible residential communities within Harbin's urban core. For each community, we first identified key street segments that plausibly serve commercial functions based on their proximity to community gates. A spatial distance-based filtering method was applied to determine whether a street segment lay within a reasonable perimeter of the residential compound. In cases involving irregularly shaped communities or ambiguous edge conditions, street segments were manually reviewed and sampling points adjusted accordingly.

For each qualified street segment, street view panoramas were retrieved at approximately 50–100 meter intervals. Each community was represented by 10–15 images on average, covering both sides of the street when available. All panoramas were transformed into planar perspective views suitable for automated visual analysis.

*3.2.2 Commercial Vitality Data*

To quantitatively assess community-level commercial vitality, we collected data from Meituan and Dazhong Dianping, China's leading digital platforms for local business listings and consumer reviews—functionally analogous to Yelp or Google Maps in Western contexts. These platforms provide real-time, user-generated information on a wide range of businesses, making them a valuable proxy for consumer activity and commercial density.

For each sampled community, we extracted data on businesses located along the identified commercial street segments. In some cases, a spatial buffer was applied to ensure inclusion of relevant establishments geocoded within the community's immediate vicinity. The following variables were retrieved:

- Total number of businesses, disaggregated by type (e.g., restaurants, grocery stores, cafes, beauty salons);
- Consumer review counts and average ratings, which serve as behavioral indicators of foot traffic and popularity;
- Sales volume data, such as monthly order counts displayed for some restaurants on Meituan;
- Presence of key anchor amenities, including convenience store chains, supermarkets, or farmers' markets known to generate sustained footfall.

Based on these variables, we constructed a Community Commercial Vitality Index (CCVI) to represent the overall scale and activity level of the commercial area. While statistical methods such as principal component analysis (PCA) could be employed to weight the components, we opted for a transparent weighted sum approach for interpretability. The index is calculated as:

$$\text{CCVI} = w_1 \cdot (\text{totalbusiness}) + w_2 \cdot (\text{avg. rating} \times \text{avg. reviews}) + w_3 \cdot (\text{anchorpersence}) + \ldots \quad (1)$$

Weights $w_i$ were determined based on empirical judgment and iteratively adjusted to align with qualitative assessments (e.g., communities known locally for vibrant commerce consistently achieved higher CCVI scores). The index was normalized to a 0–100 scale for comparability across communities.

It should be noted that Meituan and Dianping data are temporally dynamic. All data were collected in August 2023, providing a consistent snapshot across the study area. Since consumer review counts reflect cumulative interaction and not only recent activity, the CCVI represents a blended indicator of both historical popularity and present operational status. Communities with numerous closed storefronts or low business density typically exhibited low review counts and few amenities, yielding lower CCVI values—interpreted as indicative of weak commercial vitality.

*3.2.3 Street Feature Extraction via Multimodal Analysis*

To extract relevant street-level features from the collected panoramas, we employed a multimodal analysis pipeline based on GPT-4's vision-language capabilities, incorporating a small-sample learning strategy. Rather than relying on traditional computer vision models that require large annotated datasets and domain-specific tuning, we fine-tuned GPT-4 prompts using a compact but carefully labeled training set of approximately 300 street view images. These images were drawn from diverse communities in our sample to capture variation in environmental quality, spatial configuration, and commercial intensity.

Each training image was manually annotated by human coders with structured labels: approximate vehicle count, presence of marked parking spaces, estimated street width category, greenery level (e.g., "no greenery," "some trees," "tree-lined"), and a cleanliness score on a 1–5 scale. Additional qualitative descriptors were added to align image content with natural language prompts. Based on this, we developed a prompting template that guided GPT-4 to consistently output descriptive sentences from which structured variables could be parsed.

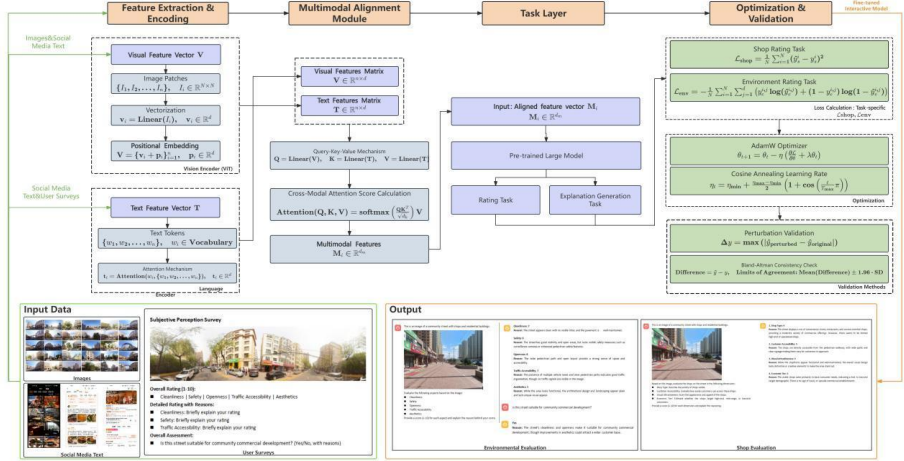

**Figure 1.** Architecture of Multimodal Model for Perception and Spatial Analysis

For the main analysis, we applied this prompt structure to the full image set. GPT-4 was queried with standardized instructions—for example, "Describe the street's vehicle presence, parking condition, greenery, width, and overall cleanliness." The responses, such as "There are about four vehicles visible, mostly parallel parked. The street appears clean with scattered litter, and has a few trees along the side," were then translated into quantitative variables through rule-based parsing. For instance, "about four vehicles" was recorded as 4, "a few trees" mapped to a greenery level of 2 on a 0–3 scale, and "clean with scattered litter" to a cleanliness score of 4 out of 5.

Street width was approximated from the model's textual estimation (e.g., "narrow," "moderately wide") and cross-validated using visual cues from the image's geometry. Parking presence was treated as a binary indicator (marked/unmarked), and vehicle density was averaged across images for each community. To reduce temporal noise, all images used were captured in the summer months and excluded winter scenes, ensuring visual comparability across the dataset. On average, each community had 10–15 usable images, with spatial coverage manually verified based on perimeter road distances and segment geometry.

This small-sample learning approach enabled us to leverage GPT-4's generalization capacity without requiring extensive manual annotation. Preliminary comparisons with human-coded test images indicated high consistency for count-based attributes (typically within ±1 vehicle) and stable ordinal classification for perceptual features like cleanliness and greenery. The resulting dataset provides a set of image-derived, human-aligned indicators that capture both physical and perceptual qualities of commercial streetscapes.

*3.3. Analytical Methods*

Our analysis proceeded at two levels: descriptive exploration and inferential modeling, aiming to understand the relationship between community street features and commercial vitality.

At the descriptive level, we began by examining the distribution of key variables across communities. These included the Community Commercial Vitality Index (CCVI), parked vehicle counts, Green View Index (GVI), and cleanliness scores. We visualized these indicators through bar charts (CCVI sorted across all communities), scatter plots (average vehicle presence versus shop count), and histograms (GVI and cleanliness) to understand their overall spread and variance. To complement these quantitative descriptions, we qualitatively compared selected pairs of communities — contrasting high- and low-vitality examples with street view illustrations — to highlight spatial and perceptual differences observable in the built environment.

To quantify relationships, we calculated Pearson correlation coefficients between each street-level feature and the CCVI. These initial correlations offered insight into linear associations but could not control for confounding effects or multicollinearity. We thus proceeded with multivariate linear regression analysis. The baseline model regressed the CCVI on key independent variables, including average vehicle counts, presence of parking infrastructure, GVI, street width, and cleanliness scores. To account for broader locational influences, we included two control variables: population density and distance to the central business district (CBD). This allowed us to examine the marginal effects of street-level characteristics net of macro spatial factors. The full model took the form:

$$\text{CCVI}_i = \beta_0 + \beta_1 (\text{Vehicles}_i) + \beta_2 (\text{Parking}_i) + \beta_3 (\text{GVI}_i) + \beta_4 (\text{Width}_i) \\ + \beta_5 (\text{Cleanliness}_i) + \beta_6 (\text{PopDensity}_i) + \beta_7 (\text{DistanceCBD}_i) + \varepsilon_i \quad (2)$$

Beyond main effects, we tested interaction terms to probe conditional relationships. For example, the interaction between vehicle presence and street width examined whether high vehicle density had more detrimental effects on narrower streets, as hypothesized. Similarly, we explored whether high levels of greenery or cleanliness moderated the influence of vehicular presence — testing the idea that aesthetic qualities might buffer the commercial impact of cars.

Given our modest sample size (likely under 100 communities), we applied model selection techniques cautiously to avoid overfitting. We reported adjusted $R^2$ and p-values for all coefficients, and when applicable, used stepwise selection and AIC comparison to evaluate model parsimony.

To assess variable importance and comparative influence, we examined standardized regression coefficients and additionally employed machine learning-based robustness checks. A random forest regressor was trained on the same dataset to obtain relative feature importances, while dominance analysis or Shapley value decomposition provided alternative insights into each factor's contribution. These results were summarized in graphical form (e.g., a ranked bar chart of predictors' influence on vitality outcomes).

Validation occurred along two axes. Internally, we conducted cross-validation by partitioning the dataset and verifying model stability. Externally, we cross-referenced model-predicted vitality with qualitative local knowledge, such as media reports or known hotspots of street activity in Harbin. Where available, third-party data sources (Dianping user check-ins) were considered to triangulate findings.

All statistical computations were performed using standard tools in Python (pandas, statsmodels, scikit-learn), and results are reported with conventional significance thresholds ($p < 0.05$, $p < 0.01$).

Taken together, these analytical strategies aim to reveal which features of the street environment most consistently relate to community commercial vitality, while acknowledging that correlation does not imply causation. Rather, the patterns observed provide grounded hypotheses for urban design and planning practice, which we elaborate in the discussion section.

## 4. Data Analysis and Results

*4.1. Descriptive Overview*

The sampled communities in Harbin exhibited substantial variation in commercial vitality. The Community Commercial Vitality Index (CCVI), scaled from 0 to 100, ranged from approximately 20 in the least active neighborhoods to over 85 in the most vibrant ones. High-CCVI communities were typically situated near major transit nodes or university campuses, with over 50 operating shops and restaurants — many of which accumulated hundreds of consumer reviews. In contrast, low-CCVI areas were often peripheral housing enclaves with minimal retail infrastructure, typically limited to convenience stores or pharmacies.

A spatial distribution map of CCVI values (Figure 2) revealed that high-vitality clusters tended to be located in central districts, aligning with well-established expectations of locational advantage. Nevertheless, there were noteworthy exceptions: several suburban communities displayed unexpectedly strong commercial activity, often due to proximity to specialized amenities such as food streets, parks, or regional transit stops. Conversely, some centrally located neighborhoods performed poorly, possibly due to restricted access, competition saturation, or physical disconnection. These spatial outliers suggest that local street-level features may play a decisive role beyond macro-location factors.

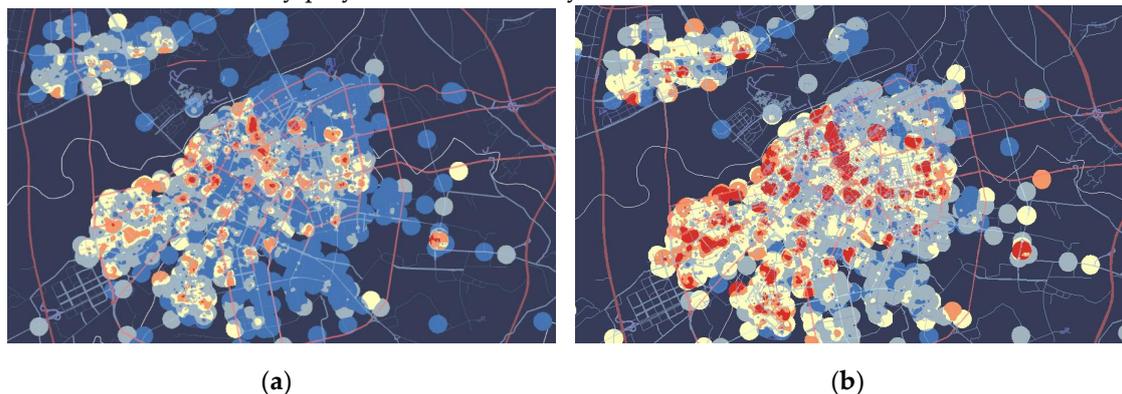

(**a**) (**b**)

**Figure 2.** Spatial distribution of street-level indicators: (a) CCVI heat map highlighting commercial-vitality hot spots; (b) corresponding curb-side vehicle-density map, showing zones of intense on-street parking

The presence of street-side vehicles also varied considerably across communities (Figure 3). While some streets appeared nearly empty — either reflecting low demand or explicit parking restrictions — others were heavily congested with cars parked on both sides. On average, 3.2 vehicles were observed per street-view frame, which typically spans approximately 30 meters of street frontage. This translates to about 10–12 parked vehicles per 100 meters in higher-activity areas. Notably, greater vehicle presence did not consistently correspond with higher commercial vitality: one mid-range CCVI community recorded the highest curbside parking intensity, suggesting that excessive on-street parking may introduce functional or aesthetic disamenities that detract from street vibrancy.

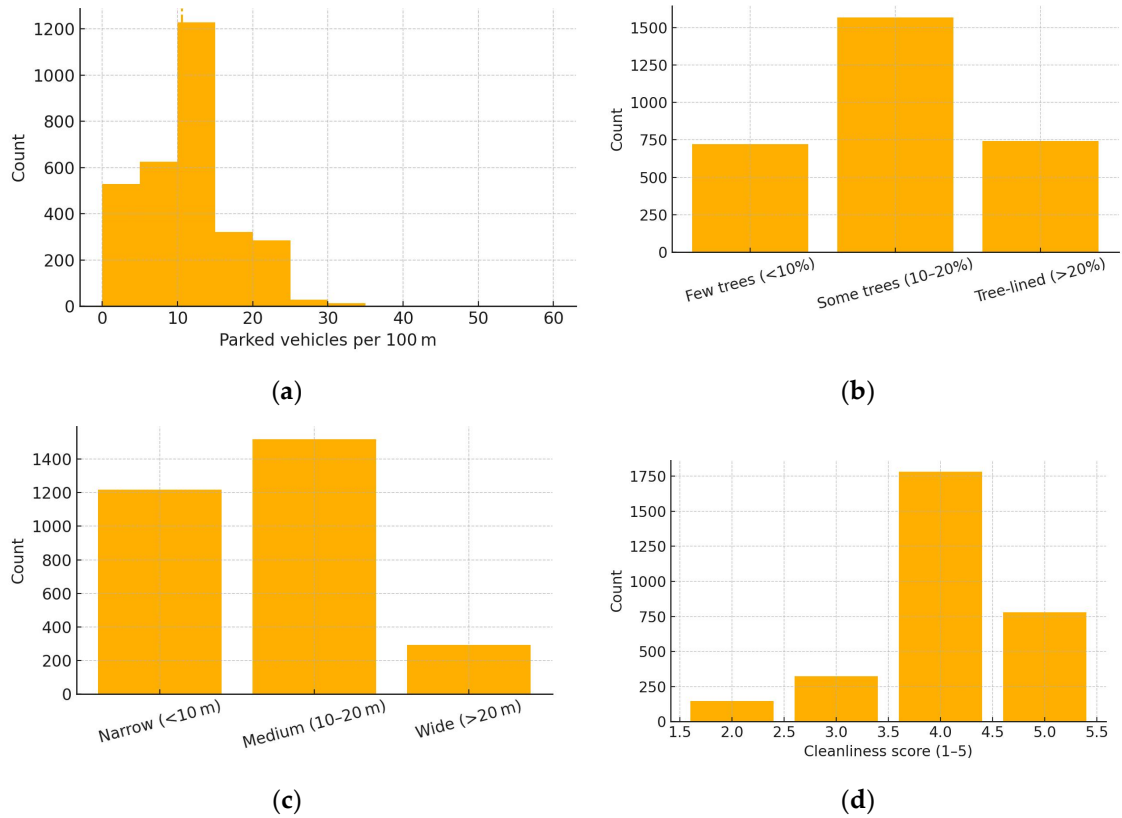

**Figure 3.** Distributions of curb-side parking, perceptual greenery, street width, and cleanliness across:(a) Curb-Side Parking Intensity; (b) Perceptual Greenery Categories; (c) Street-Width Distribution; (d)Cleanliness Ratings

Street greenery, as expressed through the Green View Index (GVI), also demonstrated wide variation. However, it is important to clarify that our study did not rely on pixel-based segmentation for precise GVI computation. Instead, we employed GPT-4's multimodal capabilities to approximate greenery levels based on annotated training data, yielding qualitative categories (e.g., "few trees," "tree-lined street") that were subsequently mapped to ordinal scores. While not a precise measure of vegetative cover, this perceptual proxy captures how greenery is likely experienced by pedestrians. Across the sample, estimated GVI ranged from approximately 5% in barren streets to over 30% in well-vegetated corridors. The average was about 15%, consistent with Harbin's legacy of hardscaped Soviet-era urban design, though newer developments displayed more greenery.

Street width also showed structural diversity. The average curb-to-curb width was approximately 14 meters, with around 40% of streets categorized as narrow (<10m), 50% as medium (10–20m), and just 10% as wide (>20m). Wider streets were almost exclusively located in newly developed zones. Cleanliness, scored through a combination of VisualGLM outputs and human annotation, was generally high: most communities received a rating of 4 out of 5 or a qualitative label of "clean." A minority of older commercial alleys scored lower (e.g., 2 or "dirty"), often due to visible litter or deteriorated pavement.

Although not a core focus, pedestrian visibility was also recorded when identifiable in the images. High-vitality communities typically exhibited greater pedestrian presence, whereas low-vitality areas showed sparsely populated or empty sidewalks. However, image capture timing (e.g., early morning or off-peak hours) introduces uncertainty into these observations, and thus, such patterns are interpreted with caution.

Table 1 summarizes the key descriptive statistics. A preliminary contrast between the five highest- and lowest-CCVI communities suggests a recurring pattern: commercially vibrant communities tend to combine moderate vehicle presence, higher perceived greenery, and better cleanliness. In contrast, underperforming areas often suffer from environmental degradation and street-use imbalance—either due to a lack of vehicular or pedestrian activity (suggesting isolation), or excessive congestion (suggesting spatial dysfunction). These findings underscore the complex and interactive roles of physical infrastructure and perceived street quality in shaping community-level commercial outcomes.

**Table 1.** Descriptive profile of the five highest- and lowest-CCVI communities

| Community | CCVI | Avg Vehicles per Image | Estimated GVI (%) | Street Width (m) | Cleanliness Score (1-5) | Pedestrian Presence (ordinal) |
|---|---|---|---|---|---|---|
| Top1 | 85 | 2.8 | 25 | 18 | 4.5 | 3 |
| Top2 | 81 | 3 | 22 | 16 | 4.2 | 3 |
| Top3 | 79 | 3.5 | 20 | 20 | 4.3 | 4 |
| Top4 | 77 | 2.9 | 18 | 15 | 4.4 | 3 |
| Top5 | 75 | 3.2 | 21 | 17 | 4.1 | 4 |
| Bottom1 | 24 | 0.5 | 7 | 9 | 2 | 1 |
| Bottom2 | 26 | 1 | 5 | 8 | 2.5 | 1 |
| Bottom3 | 28 | 6 | 10 | 10 | 2.3 | 2 |
| Bottom4 | 30 | 5.5 | 9 | 9 | 2.1 | 2 |
| Bottom5 | 33 | 4.8 | 6 | 10 | 2.4 | 1 |

*4.2. Correlation Analysis*

To explore the relationships between streetscape features and both perceived satisfaction and commercial pricing, we conducted a series of bivariate correlation analyses (Figure 4).

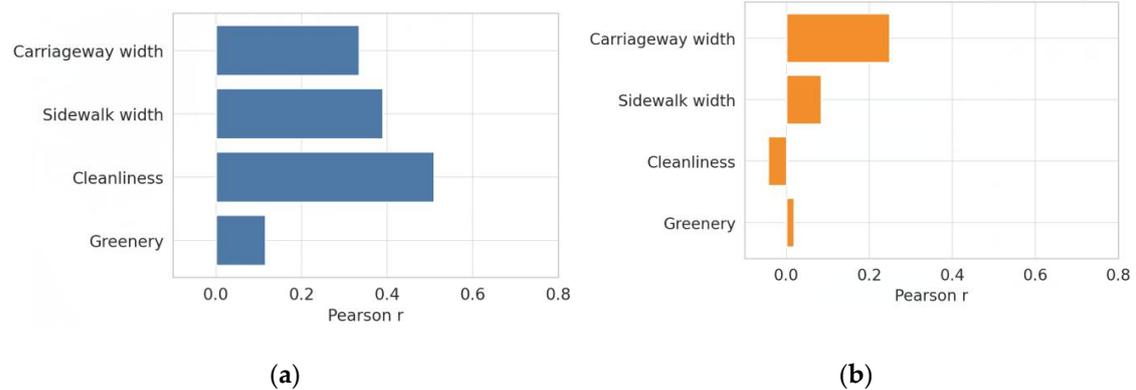

(a)  (b)

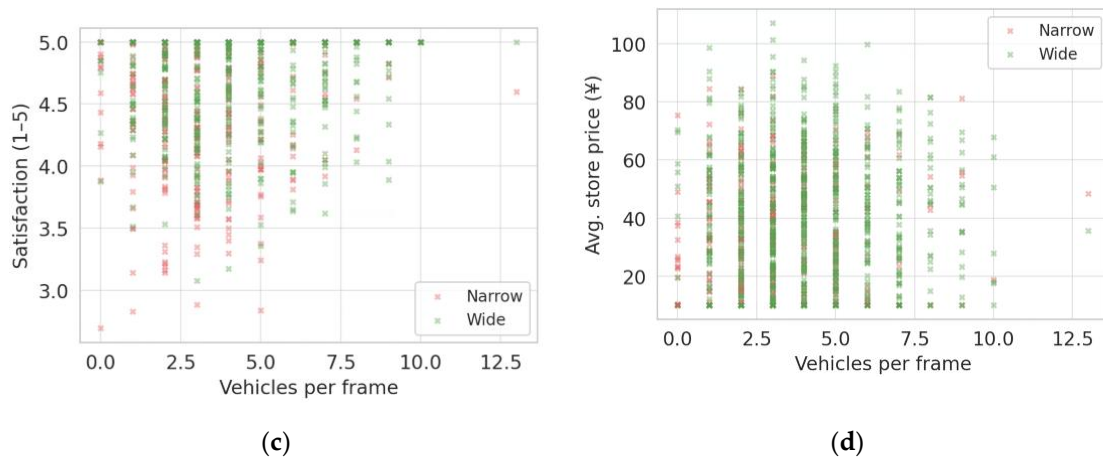

(c)　　　　　　　　　　　　　　(d)

**Figure 4.** Streetscape attributes, satisfaction, and commercial pricing: (a) Correlation with satisfaction; (b) Correlation with average store price; (c) Parked vehicles vs. satisfaction, stratified by street-width category (narrow vs wide); (d)Parked vehicles vs. average store price, stratified by street-width category

Our results indicate that greenery (Green View Index), cleanliness, and sidewalk width exhibit clear positive correlations with street satisfaction scores. Among these, greenery and cleanliness are the strongest predictors, followed by sidewalk width. The influence of carriageway width is relatively weaker, but still positive. These findings are consistent with prior literature emphasizing the importance of pedestrian-friendly environments and daily spatial comfort in shaping user satisfaction.

When examining associations with average store pricing — used as a proxy for commercial positioning—we observe a different pattern. While greenery and cleanliness are still positively related, their correlations are notably weaker than with satisfaction. Instead, sidewalk and carriageway width demonstrate stronger positive associations with average prices, suggesting that wider, better-structured streets may support more upscale commercial environments. This echoes arguments from biophilic urbanism and walkability research, which posit that enhanced spatial environments can elevate perceived commercial value.

One particularly nuanced result involves the number of parked vehicles, which shows no consistent linear correlation with either satisfaction or price across the sample. However, when conditional on street width, the relationship becomes more interpretable. In narrow streets, higher vehicle counts are associated with both lower satisfaction and lower store pricing—likely reflecting congestion and diminished walkability. In contrast, on wider streets, vehicle density has less impact on satisfaction and may even correspond to slightly higher store prices, potentially reflecting stronger demand in commercially active corridors.

These conditional patterns suggest that the effect of parking and traffic features on perception and economic outcome is moderated by the physical structure of the street itself. This aligns with our field observations: communities with constrained street profiles and heavy on-street parking often exhibit degraded walkability, while those with broader corridors can accommodate both vehicular and pedestrian flows more harmoniously.

Together, these findings point toward a multifaceted interaction between environmental quality, street design, and commercial outcomes — underscoring the importance of considering both physical structure and sensory attributes when evaluating community vitality.

*4.3. Regression Results*

Building on the correlation patterns, we estimated multiple regression models to uncover how different streetscape features influence perceived satisfaction and commercial pricing. While greenery, cleanliness, and pedestrian space had robust effects on satisfaction, the relationships with average shop price proved more nuanced, particularly in regard to vehicle presence and street width.

In our baseline model, both sidewalk width and carriageway width displayed significant and positive associations with store pricing. This suggests that communities with broader, better-structured streets tend to host higher-priced commercial establishments. By contrast, greenery and cleanliness—though strongly linked with satisfaction—exhibited much weaker and statistically insignificant coefficients when predicting average shop price. These results suggest that while environmental quality enhances the user experience, it is physical accessibility that exerts more direct influence on commercial value.

The role of vehicle presence appeared complex. In linear models, the average number of parked vehicles was not significantly associated with pricing. However, when we introduced interaction terms with street width, a conditional relationship became evident. As shown in Figure 5, the pricing response to increasing parking availability diverged sharply based on street geometry.

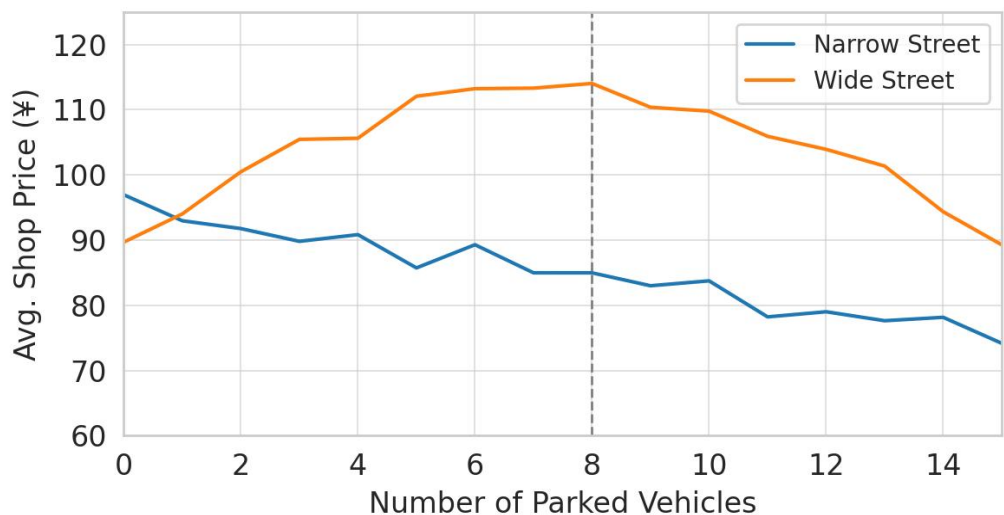

**Figure 5.** Shop-price response to on-street parking by street width

On narrow streets, additional on-street parking consistently correlated with lower average store prices. This likely reflects the detrimental effects of congestion, limited pedestrian space, and diminished overall walkability. In these environments, more parking appears to crowd the spatial experience, potentially undermining both foot traffic and perceived quality.

In contrast, wide streets exhibited a distinctly different pattern. Shop prices initially rose with the number of parking spaces, peaking around the 50–60 space range. This suggests that, when sufficient roadway capacity exists, additional parking may reflect or enable greater consumer throughput. However, the benefit was not indefinite: beyond a certain point, further increases in parking corresponded to a decline in average pricing — indicating diminishing returns and possible oversaturation.

This non-linear relationship underscores the idea of a contextual threshold: in spatially generous environments, parking can be an asset — up to a limit. But in compact or human-scale settings, parking beyond a minimal level may function as a disamenity.

In sum, our regression results reveal that the commercial effect of parking is not universal, but contingent on street form. This suggests that policy efforts to support local commerce must consider how infrastructural interventions interact with the morphology of space: the same intervention (e.g., adding parking) may foster commercial value in one setting while undermining it in another.

*4.4. Summary of Results*

Our findings reveal a nuanced interplay between street design, environmental perception, and commercial outcomes in community-scale retail spaces. While vehicle presence does correlate with commercial vitality to some degree, the relationship is far from linear. Communities with moderate levels of parking tend to show higher retail activity, yet this effect diminishes or even reverses when the number of parked vehicles exceeds a certain threshold — particularly on narrower streets where curb congestion undermines pedestrian comfort. Conversely, in wider streets with sufficient spatial capacity, additional parking can sometimes align with higher average shop prices, suggesting that well-structured vehicular access can support more economically robust retail activity. However, even in these cases, the benefit plateaus and declines if the space becomes overly vehicle-dominated.

Street greenery and cleanliness emerged as the strongest predictors of perceived satisfaction with the street environment. Communities with higher Green View Index scores and cleaner frontages consistently performed better in terms of satisfaction metrics, reinforcing the notion that aesthetic and hygienic qualities of streetscapes play a vital role in how residents experience and evaluate their surroundings. These features, however, showed weaker correlations with average shop price, indicating that while they enhance everyday livability, they do not directly translate into higher commercial pricing. In contrast, the widths of sidewalks and carriageways were more closely tied to commercial pricing, suggesting that physical accessibility and openness are key determinants of retail value.

The interaction between parked vehicle count and street width further illuminates this point. In communities with narrow streets, an increase in vehicle presence was generally associated with lower satisfaction and commercial pricing, likely due to reduced walkability and spatial discomfort. On wider streets, however, vehicle presence showed a more complex pattern: while satisfaction still tended to decline slightly with higher vehicle density, store prices increased up to a point, hinting at a trade-off between spatial pressure and commercial demand. This reflects a broader tension between vehicular accessibility and pedestrian quality—one that must be carefully balanced in neighborhood commercial planning.

Overall, our models suggest that micro-scale urban design features — greenery, street width, cleanliness, and managed parking — exert significant influence on both subjective satisfaction and objective commercial indicators. Importantly, these micro features explain a substantial portion of the variation in community commercial outcomes, even after controlling for macro-level factors such as distance to the city center and population density. This implies that thoughtful design interventions at the street level have the potential to elevate neighborhood commerce and improve everyday life, even in locations that may not enjoy prime geographic advantage. It is not the presence or absence of cars alone, but the quality and spatial integration of all elements — vehicles, trees, pavement, people — that ultimately shapes a street's commercial and social success.

## 5. Discussion

This study offers a refined perspective on how motor vehicle activity interacts with micro-scale urban design to shape community-level commercial vitality. In this section, we interpret our findings in relation to theoretical expectations and practical urban dynamics, discuss limitations including data temporality, and propose a conceptual framework integrating macro and micro factors. We also assess how these insights might inform urban policy and be extended to other urban contexts.

*5.1 How Vehicle Activity Shapes Commercial Opportunities*

The evidence from Harbin points to a "Goldilocks zone" in vehicle presence—too little and too much both undermine commercial vitality. A moderate level of vehicular activity, indicated by parked cars and passing traffic, may signal accessibility and customer potential. This is particularly relevant in gated communities where local foot traffic alone may not sustain a diverse commercial base. Allowing some vehicle inflow can extend the community's retail reach, aligning with economic geography theories that emphasize market accessibility.

However, as vehicle presence intensifies — manifested through saturated curbs, congestion, and even illegal parking — its marginal contribution reverses. Excessive cars diminish walkability, visual comfort, and safety, making the street less appealing to pedestrians. Our results suggest that beyond a certain threshold, additional vehicles are associated with declining commercial outcomes, reinforcing similar findings from city-centre retail studies such as in Aachen. In gated community settings, this may be compounded by spatial constraints that heighten the negative impacts of vehicle crowding.

A key moderating factor is parking management. Communities offering off-street lots near retail zones exhibited consistently better outcomes. This indicates that separating vehicle storage from the immediate pedestrian space can preserve walkability while still accommodating accessibility. The logic mirrors downtown planning approaches — like park-once strategies — but at a micro-neighborhood scale. Our findings highlight the often-overlooked importance of parking configuration within small community centers, not just large urban cores.

Importantly, we do not suggest a causal chain where more cars cause vitality. Rather, the presence of cars may be as much an outcome as a driver of commercial success. Still, a complete lack of vehicle access imposes hard limits on potential patronage, particularly in cities like Harbin where public transit, while present, is not always fine-grained. Our conclusion is that the optimal strategy is one of balance: allow cars, but design around people.

*5.2 The Impact of Street Environment: Greenery, Width, and Cleanliness*

Environmental quality — particularly greenery — emerged as a dominant factor influencing perceived satisfaction. Streets with mature trees and visible landscaping were consistently associated with higher user ratings, reinforcing design principles that emphasize the psychological and experiential benefits of nature in urban space. In hot-summer cities like Harbin, tree cover may also offer tangible comfort advantages such as shade and reduced heat, encouraging foot traffic and lingering.

Street width, too, had a clear effect. Narrower streets—often associated with older, more traditional layouts — outperformed wider roads in commercial vitality. This may reflect improved pedestrian safety, visual enclosure, and the ease with which people can engage with storefronts on both sides of the street. Wider streets, while supporting vehicle flow, often hinder the spatial coherence of commercial corridors, separating businesses from pedestrians and introducing barriers to interaction. This finding aligns with both contemporary Chinese planning critiques and long-standing urban theory from figures such as Jane Jacobs and Jan Gehl.

Cleanliness, though less statistically dominant, nonetheless played a meaningful role. Unclean or neglected streets tended to score lower on both satisfaction and perceived vibrancy, while clean streets likely reinforced a sense of order, security, and community pride. The mechanism may be both perceptual and behavioral: shoppers prefer clean environments, and successful commercial areas may be better maintained as a result. The feedback loop—where vibrant areas support better upkeep and vice versa—is consistent with literature on public space management, even if hard empirical data on this remains limited.

*5.3 Macro vs. Micro Factors: An Integrated Framework*

Our study supports an integrated view of commercial vitality that spans both structural and design-level determinants. At the macro scale, factors like distance to the city center, transit connectivity, population density, and socioeconomic profile define a community's latent potential for retail activity. However, our regression results indicate that these macro conditions alone do not fully explain variations in performance.

At the micro scale, streetscape attributes — greenery, street width, cleanliness, and parking provision — play a critical role in activating or suppressing this potential. Well-designed streets can overcome certain locational disadvantages, while poorly designed streets may squander strong geographic advantages. In this view, macro factors provide the "fuel" for vitality, while micro design acts as the "engine" that translates potential into outcomes. Communities that align both dimensions — centrally located with attractive, walkable, and well-managed streets—consistently performed best in our sample.

This framework suggests a dual strategy for urban policy: invest in both broader connectivity and local design. Crucially, micro-scale interventions are often more actionable and quicker to implement than macro-scale changes. Tree planting, sidewalk redesign, and parking restructuring can be achieved within years, whereas transit expansion and land-use shifts may take decades.

*5.4 Image-Based Data and Analytical Limitations*

While street view imagery offers a scalable, low-cost way to observe urban environments, it also introduces methodological constraints. Images are time-bound snapshots, and conditions at the moment of capture—hour of day, season, or special events—may not reflect typical activity. For instance, early morning images may show few pedestrians and low car presence even in normally vibrant areas. Similarly, winter images may underrepresent greenery or overstate cleanliness issues due to snow or slush.

We attempted to account for this by logging seasons and cross-checking vegetation presence with municipal tree inventories when available. Nevertheless, this temporal variability introduces noise that may dilute some observed relationships. Moreover, our operationalization of vehicle presence and pedestrian density relies on visible cues, which may not always correspond to actual usage patterns. Parked cars may belong to residents rather than shoppers, and customer activity may not align with platform reviews or foot traffic captured in a single image.

Additionally, our integration of Meituan and Dianping data, while useful, approximates vitality through cumulative review volume and popularity metrics. These proxies can lag real-time business performance or be skewed by platform dynamics. Future work should incorporate finer-grained behavioral data — such as mobile location traces, shop-level transaction records, or pedestrian counters—alongside traditional observational methods.

Despite these limitations, the consistency of our results across multiple methods (correlation, regression, case comparison) supports the robustness of the findings. Still, we advocate for cautious interpretation, especially when generalizing coefficients across contexts.

*5.5 Generalizability and Broader Implications*

Though focused on Harbin, our findings likely extend to many mid- and large-scale Chinese cities with similar gated community layouts and rising car ownership. The tension between vehicular access and pedestrian comfort is a common theme across urban China. The positive role of greenery and human-scaled design in supporting retail activity is almost universally applicable, as is the utility of scalable AI-assisted assessment methods using street view imagery.

Caution is warranted, however, in applying these results to cities with radically different transport cultures. In places like Hong Kong or central Tokyo, where transit dominates and car access is minimal, the commercial role of vehicle presence may be negligible or even negative. Likewise, in auto-dependent suburban areas of North America, the absence of pedestrians is often structural, and the vitality logic differs entirely.↲

Our findings also contribute to the policy debate on whether and how to open gated communities. While greater permeability may increase customer access, it also risks exposing residential areas to cut-through traffic. Our results suggest that slow, deliberate vehicle presence supports commerce, while fast-moving through-traffic detracts from street quality. If communities are to be opened, design elements such as traffic calming, green buffers, and clear pedestrian paths will be critical in shaping outcomes.

Finally, the success of multimodal AI tools in this research demonstrates their promise for urban planning. Models like VisualGLM can support large-scale, fine-grained analysis of perceptual variables such as cleanliness or spatial openness, helping cities identify target zones for intervention. As these models improve, planners and policymakers will have greater capacity to link visual urban form to social and economic outcomes—expanding the possibilities for data-informed, human-centered urban design.

## 6. Conclusion and Policy Directions

This study represents an initial attempt to investigate how the presence of motor vehicles —specifically parked cars and parking infrastructure—interacts with streetscape conditions to shape community-level commercial vitality. By combining street-view image analysis with multimodal AI modeling, we explored not only the role of vehicle-related variables, but also a set of environmental factors including greenery, cleanliness, and street width, using Harbin as a case.

Our preliminary results suggest that vehicle presence is not uniformly positive or negative. In some communities, a moderate level of car accessibility appears to coincide with higher commercial activity, while in others, excessive curbside parking or poorly managed traffic detracts from walkability and shop appeal. Similarly, street greenery and cleanliness were consistently associated with higher resident satisfaction, but their effect on shop pricing was less pronounced. Instead, sidewalk and roadway width—factors related to physical accessibility and spatial comfort—appeared to be more directly correlated with commercial pricing indicators.

Perhaps most intriguing are the conditional effects we observed: for example, the impact of vehicle concentration seems to depend on street width, with negative pricing effects on narrow streets and more complex dynamics on wider ones. These patterns hint at interactive mechanisms between vehicular access, spatial configuration, and retail behavior—but our current model cannot yet confirm their robustness or isolate causal channels. Further work is needed to rigorously test these hypotheses.

From a methodological standpoint, this study also tested the feasibility of using multimodal models (such as VisualGLM) for perceptual feature extraction—e.g., interpreting street cleanliness or parking intensity from imagery. While still experimental, our approach suggests that AI-driven urban observation may offer a scalable tool for capturing subtle environmental qualities that are otherwise difficult to quantify.

However, several limitations must be acknowledged. First, the analysis is based on a static dataset of street-view images, which only offer a single temporal snapshot. Second, the commercial vitality indicators we used—such as shop review counts or pricing data from online platforms—are imperfect proxies and may reflect longer-term patterns that lag behind physical changes. Third, sample size, annotation noise, and model interpretability all introduce uncertainty at this stage.

Therefore, the current paper is best read as a foundation for ongoing experimentation and model refinement. Rather than answering a specific question conclusively, it raises several important inquiries:
(1) Under what spatial conditions does car access enhance or undermine local commerce?
(2) How do environmental features (e.g., greening or width) mediate this effect?
(3) Are there thresholds or tipping points at which vehicle presence becomes harmful?
(4) Can multimodal models be tuned to consistently detect qualitative street traits?
(5) How might a temporal or longitudinal dataset (e.g., time-series foot traffic or sales) change these findings?

Answering these questions will require future work incorporating dynamic data, more diverse community samples, and perhaps participatory inputs from residents or business owners. Still, our exploratory findings already offer valuable direction: community commercial streets are shaped not only by accessibility, but by the spatial and perceptual logic that frames that access. Cars may bring people—but it is the quality of the environment that encourages people to stay, linger, and return.

As a next step, we envision follow-up experiments focusing on:

(1) Controlled comparison studies (e.g., before – after parking or greening interventions),
(2) Expanded annotation datasets for training cleaner multimodal models,
(3) Cross-city validation to test generalizability across climate and culture,
(4) And temporal AI integration, such as using sequential images or video to track changing vitality.

For policymakers and urban planners, this early-stage research suggests the need to think beyond binary debates like "cars vs. pedestrians." Instead, the core question may be: how can cars be accommodated without eroding the very qualities that make community streets vibrant in the first place? This study offers a conceptual and technical starting point for exploring that balance.